\documentclass[journal]{IEEEtran}
\usepackage{amsmath,amsfonts}
\usepackage{algorithmic}
\usepackage{algorithm}
\usepackage{array}
\usepackage[caption=false,font=normalsize,labelfont=sf,textfont=sf]{subfig}
\usepackage{textcomp}
\usepackage{stfloats}
\usepackage{url}
\usepackage{verbatim}
\usepackage{graphicx}
\usepackage{cite}
\usepackage{booktabs,pifont}
\usepackage[table]{xcolor}
\newcommand{\affmark}[1]{\raisebox{0.62ex}{\scriptsize #1}}

\begin{document}

\title{EVF-Bench: A Comprehensive Benchmark for E-commerce Visual Effects in Video Generation}

\author{Run Ling\affmark{1,*}, Ke Cao\affmark{1,*}, Runze He\affmark{2,*}, Jie Gao\affmark{1}, Ao Ma\affmark{1,{\dag},{\ddag}},\\
Wei Feng\affmark{1}, Zhanjie Zhang\affmark{1}, Hao Wang\affmark{1}, Jinghe Hu\affmark{1}, and Guibing Guo\affmark{3,{\dag}}\\[7pt]
\normalfont\normalsize
\affmark{1}\hspace{0.28em}JD.com, Beijing, China\\
\affmark{2}\hspace{0.28em}University of Chinese Academy of Sciences, Beijing, China\\
\affmark{3}\hspace{0.28em}Software College, Northeastern University, Shenyang, China\\[4pt]
E-mail: runling001@hotmail.com, maaoaoma@126.com, guogb@swc.neu.edu.cn%
\thanks{\affmark{*}\hspace{0.35em}These authors contributed equally to this work.}%
\thanks{\affmark{\dag}\hspace{0.35em}These authors are the corresponding authors.}%
\thanks{\affmark{\ddag}\hspace{0.35em}Project leader.}}

\makeatletter
\def\ps@evfpreprint{%
  \let\@oddhead\@empty \let\@evenhead\@empty
  \def\@oddfoot{\hfil\thepage\hfil}\def\@evenfoot{\hfil\thepage\hfil}}
\makeatother
\pagestyle{evfpreprint}


\maketitle
\thispagestyle{evfpreprint}

\begin{abstract}
E-commerce visual effects (EVF) are dynamic, product-centered visual presentations in e-commerce videos, often accompanied by promotional text. They play an important role in attracting viewer attention and conveying product information. However, general measures of visual quality and text relevance alone do not establish whether generated videos preserve product appearance, maintain readable text, and produce coherent, contextually appropriate effects.
To address this issue, we introduce EVF-Bench, a dedicated benchmark spanning six evaluation dimensions: visual quality and text relevance as general dimensions, and motion, product consistency, text stability, and creative rationality as EVF-specific dimensions. To complement the benchmark and support task-aligned research and model adaptation, we further construct EVF-32K, a dataset of over $32{,}000$ curated image–text–video triples tailored to e-commerce scenarios. Using this suite, we systematically evaluate representative video generation models and their domain-adapted variants. The results reveal complementary strengths, with no evaluated configuration leading all six dimensions. Together, EVF-Bench and EVF-32K provide a task-oriented evaluation framework and a supporting data resource for e-commerce video generation.
\end{abstract}


\section{Introduction}
\label{sec:introduction}

\begin{figure}[t]
  \centering
  \includegraphics[pagebox=cropbox,width=\linewidth]{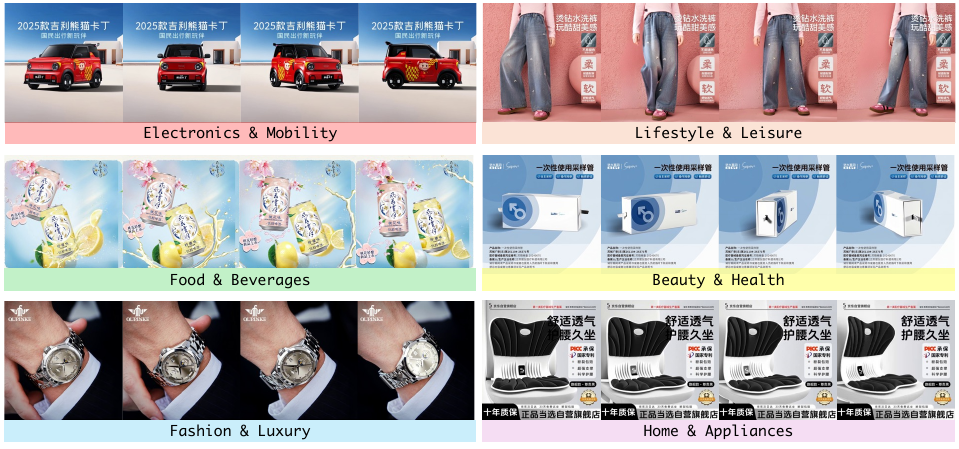}
  \caption{Examples of EVF videos across six product domains. Each group shows a sequence of sampled frames from a product-centered presentation, illustrating changes in product pose or surrounding effects alongside promotional text.}
  \label{fig:intro_evf_video_examples}
\end{figure}

\IEEEPARstart{E}{-commerce} visual effects (EVF) are dynamic, product-centered visual presentations in e-commerce videos, often accompanied by promotional text. They combine visual changes with product information to attract viewer attention and communicate the appearance and characteristics of an advertised item. Recent progress in video generation~\cite{wan2025wan,hunyuanvideo} offers a means of producing such presentations from product reference images and natural-language prompts. In the image-to-video setting studied here, the reference image provides the appearance of a specific product, while the prompt describes the desired visual changes, such as product rotation, changing illumination, or surrounding particles. EVF generation thus concerns presenting a specified product through these changes, rather than merely generating a scene that contains a product. Fig.~\ref{fig:intro_evf_video_examples} illustrates the resulting product-centered presentations through sampled video frames, including changes in product pose and surrounding effects alongside promotional text.

Unlike general-purpose video generation, EVF requires dynamic presentation and reliable product information to coexist. A visually attractive clip can still alter the product's shape or distinguishing details, make promotional text unreadable, or introduce effects that are incompatible with the item being presented. Conversely, a nearly static output may preserve appearance and text while failing to deliver the requested dynamics. These challenges motivate four EVF-specific requirements: \emph{motion}, with attention to the amplitude, smoothness, and persistence of dynamic activity; \emph{product consistency}, preserving the reference product's recognizable appearance; \emph{text stability}, maintaining clear and temporally stable promotional text when present; and \emph{creative rationality}, combining imaginative effects with coherent product presentation. The latter does not require strict realism: stylized effects can be appropriate when they remain compatible with the product and its context.

Existing benchmarks, including VBench, VBench++, EvalCrafter, and OpenS2V-Nexus, provide useful measurements of visual quality, semantic relevance, subject consistency, and temporal behavior~\cite{Vbench,vbench++,evalcrafter,yuan2025opens2vnexusdetailedbenchmarkmillionscale}. 
However, general measures of visual quality and text relevance alone do not establish whether an EVF video meets these product-centered requirements.
Whole-image similarity can be influenced by the background even when the target product changes, while prompt--video semantic agreement does not assess the readability or stability of text rendered in the frames.
Basic motion measurements also require a temporal perspective to distinguish sustained activity from isolated bursts, and visual appeal alone does not establish compatibility between creative effects and the product context.
The evaluation gap is therefore one of task alignment: related measurements exist, but EVF calls for a dedicated protocol that brings these requirements together for systematic diagnosis. Table~\ref{tab:benchmark_comparison} compares existing benchmarks with EVF-Bench across the six evaluation dimensions, indicating their coverage and suitability for EVF evaluation.

\begin{table*}[t]
  \definecolor{evfbenchmarkgreen}{RGB}{51,102,0}
  \definecolor{evfbenchmarkred}{RGB}{204,0,0}
  \definecolor{evfbenchmarkblue}{RGB}{218,232,252}
  \newcommand{\evfbenchmarkcheck}{\textcolor{evfbenchmarkgreen}{\ding{51}}}
  \newcommand{\evfbenchmarkpartial}{\textcolor{evfbenchmarkgreen}{$\triangle$}}
  \newcommand{\evfbenchmarkcross}{\textcolor{evfbenchmarkred}{\ding{55}}}
  \caption{Comparison of EVF-Bench with existing benchmarks across six evaluation dimensions. \evfbenchmarkcheck\ denotes assessment of the dimension; \evfbenchmarkpartial\ denotes related assessment that is not fully tailored to EVF; \evfbenchmarkcross\ denotes no dedicated assessment of the corresponding EVF requirement, rather than the absence of all related general-purpose measurements.}
  \label{tab:benchmark_comparison}
  \centering\small
  \setlength{\tabcolsep}{5pt}
  \renewcommand{\arraystretch}{1.08}
  \begin{tabular*}{0.95\textwidth}{@{\extracolsep{\fill}}lcccccc@{}}
    \toprule
    \textbf{Benchmark} & \shortstack{\textbf{Visual}\\\textbf{quality}} & \shortstack{\textbf{Text}\\\textbf{relevance}} & \textbf{Motion} & \shortstack{\textbf{Product}\\\textbf{consistency}} & \shortstack{\textbf{Text}\\\textbf{stability}} & \shortstack{\textbf{Creative}\\\textbf{rationality}} \\
    \midrule
    Make-a-Video-Eval~\cite{Make-a-video} & \evfbenchmarkcheck & \evfbenchmarkcheck & \evfbenchmarkcross & \evfbenchmarkcross & \evfbenchmarkcross & \evfbenchmarkcross \\
    FETV~\cite{FETV} & \evfbenchmarkcheck & \evfbenchmarkcheck & \evfbenchmarkpartial & \evfbenchmarkcross & \evfbenchmarkcross & \evfbenchmarkcross \\
    T2VScore~\cite{T2VScore} & \evfbenchmarkcheck & \evfbenchmarkcheck & \evfbenchmarkpartial & \evfbenchmarkcross & \evfbenchmarkcross & \evfbenchmarkcross \\
    EvalCrafter~\cite{evalcrafter} & \evfbenchmarkcheck & \evfbenchmarkcheck & \evfbenchmarkpartial & \evfbenchmarkcross & \evfbenchmarkcross & \evfbenchmarkcross \\
    VBench~\cite{Vbench} & \evfbenchmarkcheck & \evfbenchmarkcheck & \evfbenchmarkpartial & \evfbenchmarkcross & \evfbenchmarkcross & \evfbenchmarkcross \\
    VBench++~\cite{vbench++} & \evfbenchmarkcheck & \evfbenchmarkcheck & \evfbenchmarkpartial & \evfbenchmarkcross & \evfbenchmarkcross & \evfbenchmarkcross \\
    A2 Bench~\cite{skyreelsa2} & \evfbenchmarkcheck & \evfbenchmarkcheck & \evfbenchmarkpartial & \evfbenchmarkpartial & \evfbenchmarkcross & \evfbenchmarkcross \\
    VACE-Bench~\cite{vace} & \evfbenchmarkcheck & \evfbenchmarkcheck & \evfbenchmarkpartial & \evfbenchmarkpartial & \evfbenchmarkcross & \evfbenchmarkcross \\
    OpenS2V-Eval~\cite{yuan2025opens2vnexusdetailedbenchmarkmillionscale} & \evfbenchmarkcheck & \evfbenchmarkcheck & \evfbenchmarkpartial & \evfbenchmarkpartial & \evfbenchmarkcross & \evfbenchmarkcross \\
    \midrule
    \textbf{EVF-Bench} & \evfbenchmarkcheck & \evfbenchmarkcheck & \evfbenchmarkcheck & \evfbenchmarkcheck & \evfbenchmarkcheck & \evfbenchmarkcheck \\
    \bottomrule
  \end{tabular*}
\end{table*}

\begin{table*}[h]
  \definecolor{evfdatasetblue}{RGB}{218,232,252}
  \caption{Comparison of EVF-32K with existing video datasets in terms of content focus, resolution, clip count, average clip length, and total video duration. All entries follow each dataset's own reported statistics at the video level, so that the three columns are mutually consistent; average lengths are as reported or derived from the reported duration and clip count; -- indicates a value not reported.}
  \label{tab:dataset_comparison}
  \centering\small
  \setlength{\tabcolsep}{5pt}
  \renewcommand{\arraystretch}{1.08}
  \begin{tabular*}{0.95\textwidth}{@{\extracolsep{\fill}}lccccc@{}}
    \toprule
    \textbf{Dataset} & \textbf{Scene} & \textbf{Resolution} & \shortstack{\textbf{Video}\\\textbf{clips}} & \shortstack{\textbf{Average}\\\textbf{length (s)}} & \shortstack{\textbf{Video}\\\textbf{duration (h)}} \\
    \midrule
    Panda-70M~\cite{Panda-70M} & Open & 720P & 70M & 8.6 & 167K \\
    ChronoMagic-Pro~\cite{chronomagicbench} & Time-Lapse & 720p & 460K & 234.8 & 30K \\
    OpenHumanVid~\cite{openhumanvid} & Human & 720P & 52.3M & 4.9 & 70K \\
    ActivityNet~\cite{activitynet} & Action & -- & 27.8K & 110.0 & 849 \\
    YouCook2~\cite{youcook2} & Cooking & -- & 2K & 316.0 & 176 \\
    How2~\cite{how2} & Instruct & -- & 80K & 90.0 & 2000 \\
    OGameData~\cite{che2024gamegenxinteractiveopenworldgame} & Game & 720P-4K & 1000K & -- & 4K \\
    OpenS2V-5M~\cite{yuan2025opens2vnexusdetailedbenchmarkmillionscale} & Subject & 720P & 5.4M & 6.6 & 10K \\
    \midrule
    \textbf{EVF-32K} & Product & 960$\times$960 & 32K & 5.0 & 44.4 \\
    \bottomrule
  \end{tabular*}
\end{table*}

To address this gap, we introduce \textbf{EVF-Bench}, a dedicated benchmark containing 176 evaluation cases across six product domains. The benchmark spans six evaluation dimensions: visual quality and text relevance as general dimensions, and motion, product consistency, text stability, and creative rationality as EVF-specific dimensions. 
Beyond general quality and relevance, the evaluation emphasizes fidelity to the reference product, the clarity and temporal stability of promotional text when present, and the quality and contextual appropriateness of dynamic effects. 
This design provides a dimension-wise performance profile, enabling the identification of strengths and limitations that general quality and relevance scores alone may not reveal.


To complement the benchmark and support task-aligned research and model adaptation, we further construct \textbf{EVF-32K}, a dataset of over $32{,}000$ curated image--text--video triples spanning six product domains. General video corpora such as Panda-70M and OpenVid-1M provide broad content coverage~\cite{Panda-70M,openvid1m}, but their scale alone does not ensure examples tailored to the joint requirements of product presentation, promotional text, and dynamic effects. EVF-32K connects product reference images with prompts describing desired visual changes and corresponding videos. Its construction combines product and image selection, prompt construction, image-to-video generation, video filtering, and metadata augmentation. Whereas EVF-Bench defines evaluation cases and measurements, EVF-32K provides curated examples for analyzing and adapting models to the same application setting. Table~\ref{tab:dataset_comparison} compares its content focus and reported statistics with those of existing video datasets.

Using this suite, we systematically evaluate four commercial systems, four open-source models, and two variants adapted using EVF-32K. The results reveal complementary strengths: no evaluated configuration leads all six dimensions. For example, Kling 2.1 Std obtains the highest scores in visual quality, motion, and text stability, Taobao Xingchen leads in text relevance, and Wan2.2-EVF leads in product consistency and creative rationality. Both adapted variants improve on their corresponding backbones across the six reported scores. These findings demonstrate the diagnostic value of multi-dimensional evaluation and motivate further study of task-aligned adaptation, without reducing EVF performance to a single model ranking.

The contributions of this work are summarized as follows:
\begin{itemize}
  \item \textbf{EVF-Bench.} We introduce a dedicated benchmark with 176 evaluation cases and six complementary dimensions, combining general visual quality and text relevance with EVF-specific motion, product consistency, text stability, and creative rationality.
  \item \textbf{EVF-32K.} We construct a supporting dataset of over $32{,}000$ curated image--text--video triples and a multi-stage construction pipeline for task-aligned research and model adaptation.
  \item \textbf{Empirical evaluation.} We compare representative video generation systems and their domain-adapted variants, identify complementary strengths across the evaluation dimensions, and analyze the improvements obtained through adaptation.
\end{itemize}

\section{Related Work}
\label{sec:related}

\subsection{Evaluation of Video Generation}
\label{sec:related_evaluation}
Video generation benchmarks have moved beyond aggregate quality scores toward more diagnostic evaluation. VBench~\cite{Vbench} organizes video quality into distinct dimensions, including subject consistency, motion smoothness, temporal flickering, and spatial relationships, with corresponding prompts and evaluation methods. VBench++~\cite{vbench++} extends this approach to both text-to-video and image-to-video generation, introducing an image suite for reference-conditioned assessment. EvalCrafter~\cite{evalcrafter} combines a diverse prompt suite with measurements of visual quality, content quality, motion, and text--video alignment, and calibrates its scores against human judgments. These works establish important foundations for comparing models across multiple aspects of generation quality.

Other studies emphasize fine-grained semantic evaluation and the reliability of automatic metrics. FETV~\cite{FETV} categorizes prompts by content, controllable attributes, and complexity, including temporal categories, and uses human evaluation to examine model performance and metric reliability. T2VScore~\cite{T2VScore} separately evaluates text--video alignment and video quality, supported by human judgments. The assessment of machine-generated video has also received dedicated attention, including label-free metric learning for general AI-generated video quality~\cite{zhang2026lmvq}, transformation-consistency-based quality prediction for AI-generated content videos~\cite{xu2026aigcvqa}, and a motion-centered quality benchmark for text-to-video generation~\cite{zhang2026mogenvd}. For reference-conditioned generation, OpenS2V-Nexus~\cite{yuan2025opens2vnexusdetailedbenchmarkmillionscale} introduces OpenS2V-Eval to assess subject consistency, naturalness, and text relevance. Its explicit focus on reference-subject fidelity is particularly relevant to product-centered generation. VACE-Bench~\cite{vace} further broadens the evaluation setting to reference-based generation and video editing tasks.

A growing share of these protocols relies on multimodal large language models as judges. The reliability of such judges is itself an open question: recent evaluation shows that multimodal models still struggle with mathematical spatial reasoning over two- and three-dimensional relations~\cite{lu2026domllmsspace}. This motivates rubrics that bind each rating to explicit, dimension-specific criteria instead of relying on free-form model preference.

Building on these studies, EVF-Bench focuses on the evaluation requirements of e-commerce product presentations. It combines visual quality and text relevance with four EVF-specific dimensions: motion, product consistency, text stability, and creative rationality. By examining how dynamic effects coexist with faithful product appearance and stable promotional text when present, EVF-Bench complements existing benchmarks with an evaluation protocol tailored to e-commerce video generation.

\subsection{Datasets for Visual Generation}
\label{sec:related_datasets}
Visual generation research draws on both general-purpose corpora and application-oriented resources. LAION-5B~\cite{schuhmann2022laion} provides large-scale image--text pairs, while DiffusionDB~\cite{wang2022diffusiondb} records generated images together with prompts and generation parameters. Product-oriented research further addresses the requirements of e-commerce imagery: BG60k~\cite{wang2025generate} supports product background generation with category-dependent and personalized styles, and RF1M~\cite{du2024towards} contains generated advertising images annotated with human feedback to support reliability assessment. These resources contribute to static product presentation and image-level quality control, but do not directly provide temporal observations of product appearance, rendered text, or dynamic effects.

In the video domain, WebVid~\cite{WebVid} and Panda-70M~\cite{Panda-70M} provide web-sourced video--text data, while OpenVid-1M~\cite{openvid1m} emphasizes curated videos with expressive captions for text-to-video generation. OpenS2V-5M, introduced with OpenS2V-Nexus~\cite{yuan2025opens2vnexusdetailedbenchmarkmillionscale}, is more closely related to our setting because it supplies subject--text--video triples for subject-conditioned generation. Such resources offer broad content coverage or explicit subject conditioning; however, those properties alone do not ensure examples curated for the combined requirements of product presentation, promotional text, motion, and creative rationality.

EVF-32K complements these datasets with product-centered image--text--video triples curated for the application setting evaluated by EVF-Bench. Whereas EVF-Bench specifies evaluation cases and measurements, EVF-32K supplies examples for task-aligned analysis and model adaptation. This distinction preserves the complementary roles of the evaluation benchmark and the supporting data resource. Table~\ref{tab:dataset_comparison} summarizes the content focus and reported statistics of the compared datasets.

\subsection{Controllable and Subject-Conditioned Generation}
\label{sec:related_generation}
Generators increasingly expose explicit control over layout, style, identity, and rendered text, which is what product-centered presentation requires. Within diffusion transformers, Lay2Story~\cite{ma2025lay2story} extends a transformer backbone so that multi-frame stories can be generated under layout-level toggles, whereas PlanGen~\cite{he2025plangen} unifies layout planning and image synthesis in a single auto-regressive vision--language model so that spatial arrangement is decided before pixels are produced. RelaCtrl~\cite{cao2026relactrl} instead targets the cost of control, injecting relevance-guided signals into selected transformer layers to reduce the parameter and computation overhead of conditional generation, and U-StyDiT~\cite{zhang2025ustydit} applies the same backbone family to ultra-high-resolution artistic style transfer.

For video, FancyVideo~\cite{feng2025fancyvideo} replaces per-frame textual guidance with cross-frame guidance to improve both dynamics and temporal consistency, while CustomTTT~\cite{bi2025customttt} customizes motion and appearance through test-time training rather than offline fine-tuning. MoFu~\cite{ling2026mofu} addresses multi-subject generation by combining scale-aware modulation with Fourier-domain fusion so that each referenced subject keeps a natural scale and appearance, and WISA~\cite{wang2025wisa} adds physics-aware assistance to text-to-video generation. Rendered text has been treated as a first-class target as well: InnoText~\cite{liu2026innotext} unifies visual text generation and editing in a single model. Personalization is studied in the image domain, e.g., RAGAR~\cite{ling2025ragar} renders reference images into preferred ones under the guidance of a recommendation signal. Because such controls are usually validated by aggregate fidelity or aesthetic scores on general data, cross-domain robustness has been pursued separately, for instance by routing selective mixture-of-experts with foundation-model features for stereo matching in the wild~\cite{wang2025stereomoe}.

EVF-Bench complements these generation-side advances by measuring the outcomes on which e-commerce presentation depends: whether the requested motion occurs and persists, whether the referenced product and the rendered promotional text survive the effects, and whether the effect remains compatible with the product context. Control mechanisms and adaptation strategies can therefore be compared under one protocol rather than by architecture-specific criteria.

\section{EVF-Bench}
\label{sec:benchmark}

\subsection{Benchmark Construction and Task Setting}
\label{sec:benchmark_construction}
\textbf{Task setting.}
EVF-Bench evaluates image-to-video generation for product-centered e-commerce presentations. Each evaluation case comprises a product reference image $I$, a natural-language prompt $p$, and an associated product name $c$. The image provides the visual reference for the target product, while the prompt specifies the desired visual changes. Given these generation conditions, a model $G$ produces a video
\begin{equation}
  V=G(I,p), \qquad V=\{F_t\}_{t=1}^{T},
  \label{eq:benchmark_generation_task}
\end{equation}
where $F_t$ denotes the $t$th frame. The product name supports product localization during evaluation; it is not an additional generation input in this task formulation. The generated video is evaluated for its visual quality, relevance to the prompt, and preservation of the product-centered presentation requirements. Multiple videos can satisfy the same image and prompt, so evaluation does not rely on pixel-level agreement with a unique target video.

\textbf{Case construction.}
We construct the evaluation cases by pairing e-commerce product reference images with prompts specifying a dynamic presentation. The reference images provide product appearance and scene context, often including promotional text. Each prompt describes how the product or its surroundings should change over time, such as through product movement, illumination changes, or surrounding visual effects, rather than merely restating the static image. Cases containing promotional text additionally enable evaluation of whether that information remains clear and stable in the generated video. The resulting image--prompt pairs, together with the associated product names, form the benchmark's test conditions. Benchmark construction therefore concerns the selection of reference images and specification of visual changes; the generation and filtering of video examples for model adaptation are addressed separately in the EVF-32K dataset section.

\begin{figure}[ht]
  \centering
  \includegraphics[pagebox=cropbox,width=0.95\linewidth]{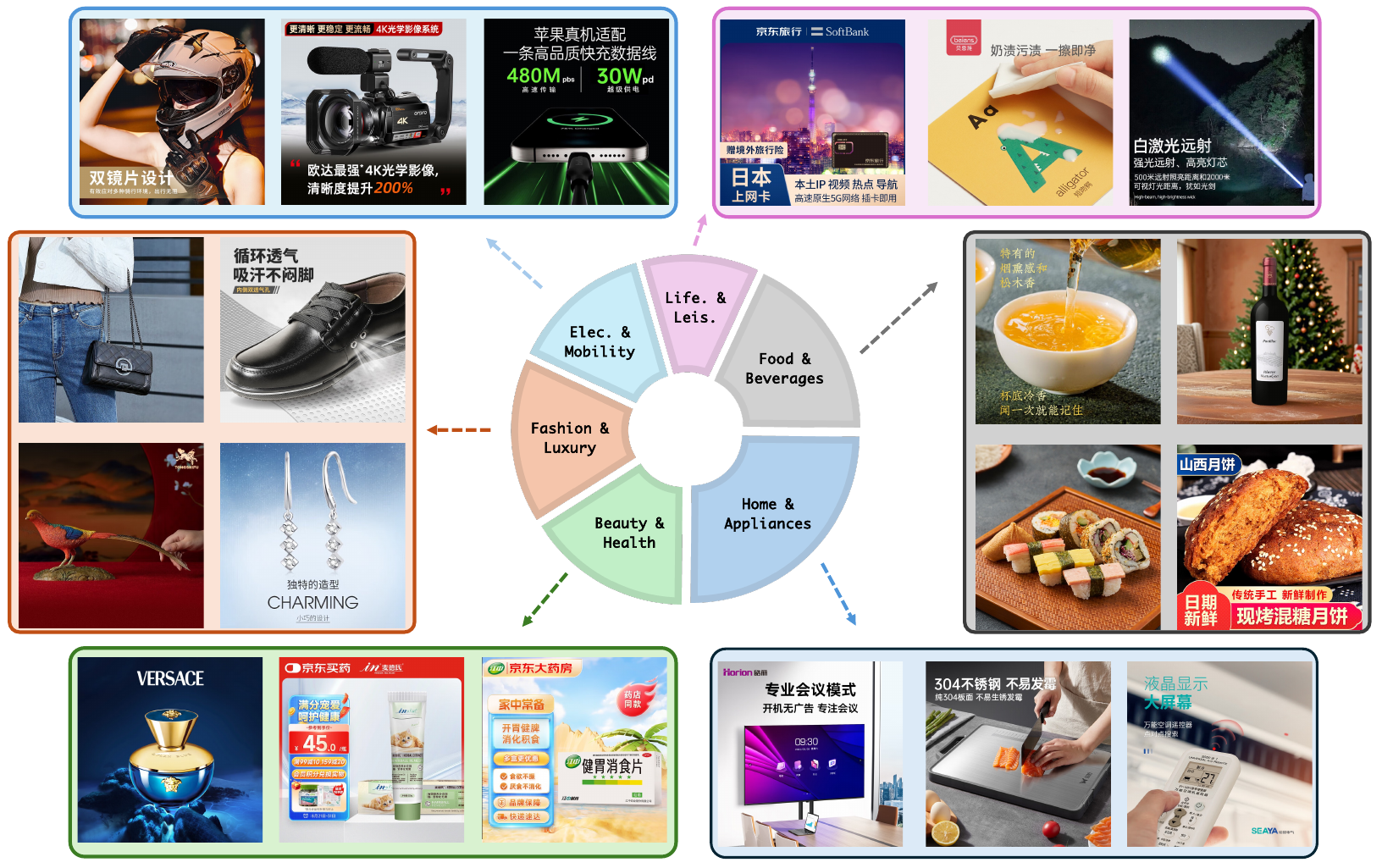}
  \caption{Representative product reference images from the six domains of EVF-Bench. The examples illustrate the variation in product appearance, promotional text, and presentation context covered by the evaluation cases.}
  \label{fig:benchmark_domain_examples}
\end{figure}

\textbf{Domain coverage.}
The benchmark contains 176 evaluation cases across six product domains: \emph{Home \& Appliances} with 44 cases (25.00\%), \emph{Food \& Beverages} with 32 (18.18\%), \emph{Beauty \& Health} and \emph{Fashion \& Luxury} with 28 each (15.91\%), \emph{Electronics \& Mobility} with 24 (13.64\%), and \emph{Lifestyle \& Leisure} with 20 (11.36\%). Fig.~\ref{fig:benchmark_domain_examples} presents representative reference images from these domains. This coverage includes different product geometries, materials, text layouts, and presentation contexts. For example, food presentations emphasize texture and color, clothing involves deformable materials, luxury items introduce reflective surfaces, and electronics and appliances expose geometric details. These variations allow evaluation across heterogeneous product presentations rather than a single object type or visual style.

\begin{figure*}[t]
  \centering
  \includegraphics[pagebox=cropbox,width=0.95\textwidth]{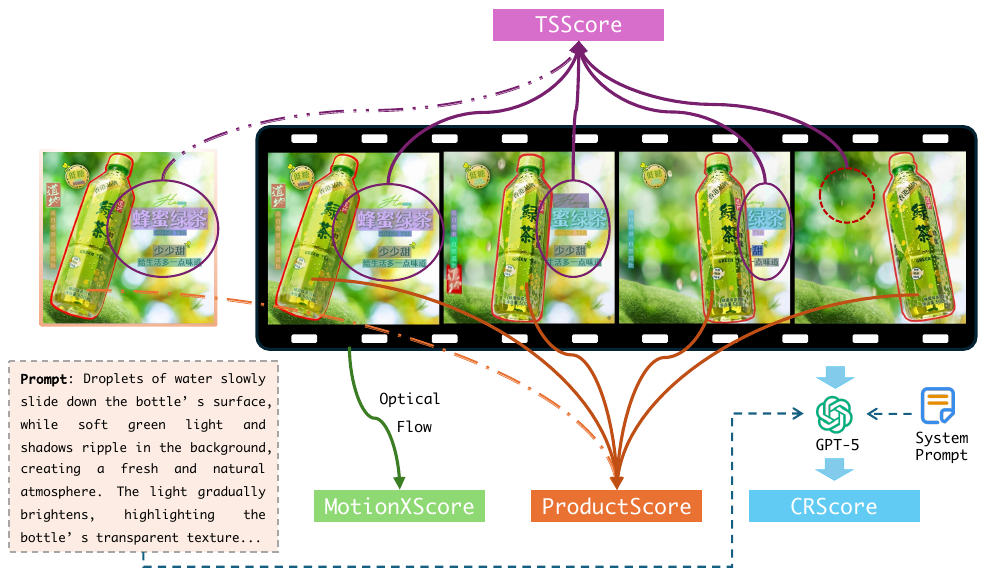}
  \caption{Design of the four EVF-specific metrics in EVF-Bench. ProductScore compares the reference product with its generated appearance; TSScore follows rendered text across sampled frames; MotionXScore characterizes motion through optical flow; and CRScore uses the video, generation prompt, and evaluation instruction to judge creative rationality. AesScore and GmeScore provide complementary assessments of visual quality and text relevance and are not shown here.}
  \label{fig:benchmark_metric_design}
\end{figure*}

\subsection{Evaluation Metrics}
\label{sec:benchmark_metrics}
EVF-Bench evaluates generated videos along six complementary dimensions. AesScore and GmeScore assess visual quality and text relevance, respectively, while ProductScore, TSScore, MotionXScore, and CRScore address product consistency, text stability, motion, and creative rationality. The first two metrics follow established general evaluation designs; the remaining four focus on the product, rendered text, dynamic activity, and contextual compatibility of EVF presentations. Fig.~\ref{fig:benchmark_metric_design} illustrates the observations used by these four EVF-specific metrics.

\subsubsection{\textbf{Visual Quality}}
\label{sec:benchmark_aesscore}
AesScore measures the aesthetic quality of generated frames, following the design adopted by OpenS2V~\cite{yuan2025opens2vnexusdetailedbenchmarkmillionscale,schuhmann2024aesthetic}. A CLIP image encoder~\cite{clip} extracts frame representations, which are passed to a multi-layer perceptron trained for aesthetic prediction. Learning perceptual quality from such representations follows the practice of automatic video quality assessment, which predicts human judgments of user-generated video with spatiotemporal models~\cite{zhu2023ugcvqa}. Let $\mathcal{F}_{\rm aes}$ denote the sampled frames and $a(F)$ the predicted aesthetic score of frame $F$. The video-level score is
\begin{equation}
  s_{\rm aes}(V)=\frac{1}{|\mathcal{F}_{\rm aes}|}
  \sum_{F\in\mathcal{F}_{\rm aes}}a(F).
  \label{eq:benchmark_aesscore}
\end{equation}
This metric provides a general assessment of visual appeal, rather than a direct test of product identity or promotional-text preservation.

\subsubsection{\textbf{Text Relevance}}
\label{sec:benchmark_gmescore}
GmeScore measures semantic relevance between the generation prompt and the generated video. In EVF, prompts describe not only the product and scene but also the desired visual changes, such as product movement, lighting transitions, and surrounding effects. 
Following OpenS2V~\cite{yuan2025opens2vnexusdetailedbenchmarkmillionscale}, we adopt GME, a multi-modal retrieval model built upon Qwen2-VL~\cite{gme}. Its embedding-based relevance assessment accommodates prompts of varying lengths, making it suitable for these descriptive instructions. The generation prompt $p$ and sampled video content provide the textual and visual inputs, respectively, with the scoring procedure inherited from OpenS2V. The resulting score summarizes the semantic correspondence between the requested presentation and the generated content.

\subsubsection{\textbf{Product Consistency}}
\label{sec:benchmark_productscore}
ProductScore assesses whether the generated video preserves the appearance of the reference product. 
It adapts NexusScore~\cite{yuan2025opens2vnexusdetailedbenchmarkmillionscale}, which localizes a reference subject in generated frames and evaluates subject--video alignment. 
E-commerce reference images often contain promotional text, background graphics, and decorative elements. Using the entire image as the visual reference can therefore introduce information unrelated to the target product.

We use product-name-guided grounding and SAM2 segmentation~\cite{liu2023grounding,ravi2024sam2segmentimages} to extract the product from the reference image. 
Let $M(I,c)$ denote the product mask obtained from image $I$ and product name $c$. The resulting crop is supplied to the NexusScore evaluator:
\begin{equation}
  \begin{aligned}
    I_{\rm prod}&=\operatorname{Crop}(I,M(I,c)),\\
    s_{\rm prod}(I,V)&=\operatorname{Nexus}(I_{\rm prod},V).
  \end{aligned}
  \label{eq:benchmark_productscore}
\end{equation}
This adaptation changes the reference preprocessing while retaining the underlying subject-matching and aggregation design. It reduces background and overlay interference without imposing pixel-level agreement, which would penalize legitimate changes in pose or illumination.

\subsubsection{\textbf{Text Stability}}
\label{sec:benchmark_tsscore}
TSScore evaluates text rendered in the video rather than the generation prompt. Rendered text has itself become an explicit target of generation and editing models~\cite{liu2026innotext}, so its legibility and persistence are assessed separately from prompt relevance. TSScore focuses on non-product text overlays by excluding the product region, reducing interference from logos and packaging text. From $N_{\rm txt}$ uniformly sampled frames, OCR~\cite{cui2025paddleocr30technicalreport} detects text boxes and recognized strings. Spatial overlap and string similarity associate detections across frames into text tracks $\mathcal{L}=\{L_j\}$. For track $L_j$, let $n_j$ be its number of observations, $u_{j,r}$ the recognized string at observation $r$, $b_{j,r}$ its text box, and $c_{j,r}$ the OCR confidence.

\emph{Content stability} measures cross-frame consistency of recognized strings. For $n_j\geq2$, it is defined as
\begin{equation}
  \operatorname{Stab}(L_j)=1-\frac{1}{n_j-1}
  \sum_{r=1}^{n_j-1}\operatorname{CER}(u_{j,r},u_{j,r+1}),
  \label{eq:benchmark_text_stability}
\end{equation}
where CER is the character error rate. This temporal self-consistency measurement does not require a ground-truth transcription and does not establish that a consistently rendered string is correct.

\emph{Clarity} combines recognition confidence with the visual sharpness of each text region:
\begin{equation}
  \begin{aligned}
    \operatorname{Clarity}(b_{j,r})={}&\lambda_c c_{j,r}\\
    &+(1-\lambda_c)\operatorname{NormVarLap}(b_{j,r}),
  \end{aligned}
  \label{eq:benchmark_text_clarity}
\end{equation}
where $\operatorname{NormVarLap}$ denotes normalized Laplacian variance. Averaging over the observations gives track-level clarity $\overline{\operatorname{Clarity}}(L_j)$. 

\emph{Temporal coverage} measures how often the tracked text is detected among the sampled frames:
\begin{equation}
  \operatorname{Cover}(L_j)=\frac{n_j}{N_{\rm txt}}.
  \label{eq:benchmark_text_coverage}
\end{equation}
The three components are combined by weighted averaging across tracks:
\begin{equation}
  \begin{aligned}
    s_{\rm text}(V)=\frac{1}{|\mathcal{L}|}
    \sum_{L_j\in\mathcal{L}}\big[&w_s\operatorname{Stab}(L_j)\\
    &+w_c\overline{\operatorname{Clarity}}(L_j)\\
    &+w_p\operatorname{Cover}(L_j)\big].
  \end{aligned}
  \label{eq:benchmark_tsscore}
\end{equation}
We use equal component weights $w_s=w_c=w_p=1/3$ to give content stability, clarity, and coverage the same contribution. The sampling count $N_{\rm txt}$ and clarity mixing coefficient $\lambda_c$ specify the evaluation configuration, which is reported in Sec.~\ref{sec:experimental_implementation}. 
TSScore combines content stability, legibility-related observations, and persistence, rather than evaluating prompt relevance or text accuracy alone.

\subsubsection{\textbf{Motion}}
\label{sec:benchmark_motionxscore}
MotionXScore extends the amplitude-and-smoothness evaluation described in OpenS2V~\cite{yuan2025opens2vnexusdetailedbenchmarkmillionscale} with an activation ratio that characterizes motion persistence. 
The additional component distinguishes activity sustained across a video from motion concentrated in short bursts. 
Let $E_t$ denote the mean optical-flow magnitude between consecutive frames $F_t$ and $F_{t+1}$, for $t=1,\ldots,T-1$. An exponential moving average reduces short-term fluctuations:
\begin{equation}
  \widetilde E_t=\alpha_m E_t+(1-\alpha_m)\widetilde E_{t-1}.
  \label{eq:benchmark_motion_smoothing}
\end{equation}
An adaptive threshold $\tau$ based on robust motion statistics identifies active intervals. To suppress isolated spikes, an interval contributes to the activation ratio only if it belongs to a contiguous active run of at least $L_{\min}$ seconds. Writing $d_t$ for the duration of that run, valid activity is indicated by
\begin{equation}
  z_t=\begin{cases}
    1,&\widetilde E_t\geq\tau\ \text{and}\ d_t\geq L_{\min},\\
    0,&\text{otherwise}.
  \end{cases}
  \label{eq:benchmark_motion_activity}
\end{equation}
The activation ratio is
\begin{equation}
  R=\frac{1}{T-1}\sum_{t=1}^{T-1}z_t.
  \label{eq:benchmark_motion_activation}
\end{equation}
The smoothing coefficient $\alpha_m$ and minimum run duration $L_{\min}$ specify the temporal evaluation configuration. Expressing run length in elapsed time separates this criterion from the video sampling rate. Finally, MotionXScore averages the normalized amplitude, smoothness, and activation-ratio components:
\begin{equation}
  s_{\rm motion}(V)=\frac{1}{3}(\widehat A+\widehat S+\widehat R).
  \label{eq:benchmark_motionxscore}
\end{equation}
Here, $\widehat A$ and $\widehat S$ represent the inherited motion components, while $\widehat R$ represents the added persistence component. Optical flow can also reflect camera movement or product deformation, so activity should be interpreted together with product consistency and creative rationality, rather than as proof that the prompt-specified effect was achieved.

\subsubsection{\textbf{Creative Rationality}}
\label{sec:benchmark_crscore}
CRScore evaluates the balance between creative visual effects and coherent product presentation. It uses a multi-modal large language model~\cite{openai20205gpt5} as a proxy for human judgment of originality, product integrity, and compatibility with the scene context. A stylized effect need not obey strict physical realism, but it should remain compatible with the product and the intended presentation.

Uniformly sampled video frames are supplied to the evaluator with a structured scoring instruction and the generation prompt. The prompt provides context rather than a literal checklist for evaluating every word. 
The instruction defines five rating anchors: 1 for little creativity with severe presentation or product-integrity problems; 2 for limited creative attempts with clear coherence problems; 3 for moderate creativity with mostly consistent products and some flaws; 4 for clear creativity and coherent presentation with minor flaws; and 5 for strong creativity together with consistent, coherent product presentation. 
The evaluator returns a discrete rating $r^{(k)}\in\{1,2,3,4,5\}$ for each repeated judgment, and the final raw score is
\begin{equation}
  s_{\rm cr}(V,p)=\frac{1}{N_{\rm cr}}
  \sum_{k=1}^{N_{\rm cr}}r^{(k)}.
  \label{eq:benchmark_crscore}
\end{equation}
The sampled frame count and number of repeated judgments are specified by the evaluation configuration. Averaging repeated judgments is intended to reduce variability, while the rubric keeps creativity and contextual coherence within the same assessment. CRScore is a perceptual proxy for creative rationality, not a measurement of advertising effectiveness or commercial success.

\section{EVF-32K Dataset}
\label{sec:dataset}
EVF-32K is a collection of over $32{,}000$ curated image--text--video triples for e-commerce visual effects. Each example associates a product reference image with a video and a textual description. The construction process uses an image-conditioned generation prompt to specify the desired presentation and adds a video caption describing the resulting content. The prompt guides generation, while the caption describes observed product appearance and dynamics. The dataset provides examples for task-aligned research and model adaptation, complementing the evaluation cases and metrics defined by EVF-Bench.

\subsection{Dataset Construction}
\label{sec:dataset_construction}
\begin{figure*}[t]
  \centering
  \includegraphics[pagebox=cropbox,width=0.95\textwidth]{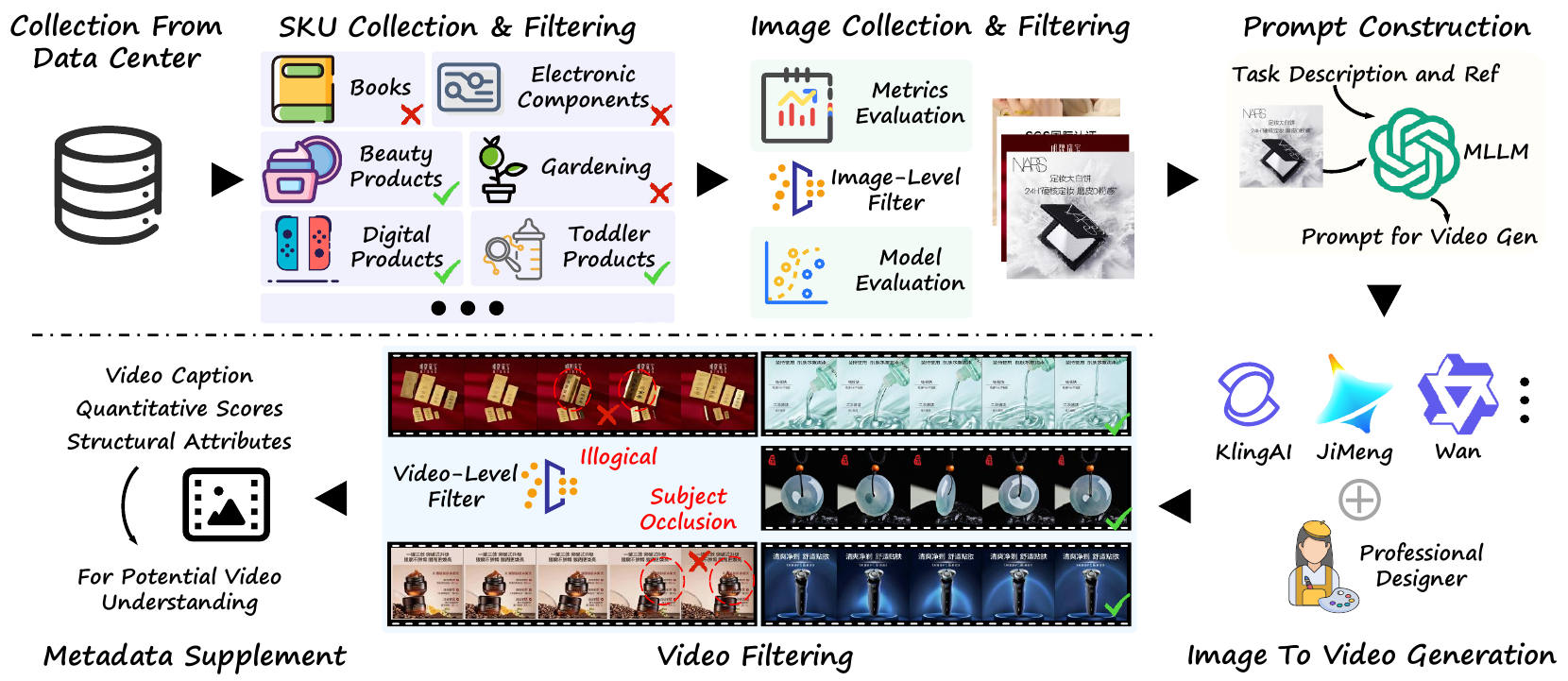}
  \caption{The six-stage construction pipeline of EVF-32K. Product and image filtering provide reference inputs for prompt construction and video generation; video filtering selects candidate clips, and metadata augmentation adds captions, quantitative indicators, and structural attributes.}
  \label{fig:dataset_construction_pipeline}
\end{figure*}

The construction pipeline comprises six stages. Fig.~\ref{fig:dataset_construction_pipeline} summarizes how product references are converted into video examples with associated descriptions and auxiliary information.

\subsubsection{\textbf{SKU Collection and Filtering}}
We collect product categories across major e-commerce domains and select those suitable for product-centered visual presentation. Selection considers whether products have identifiable visual structure and support observable changes in the product or its surroundings. Categories without a tangible visual representation, or with limited relevance to dynamic product presentation, are excluded. This stage establishes the product coverage for subsequent image collection.

\subsubsection{\textbf{Image Collection and Filtering}}
For each selected category, product images are collected as references for prompt construction and video generation. Filtering considers four aspects: image quality, layout structure, subject saliency, and text interference. 
Laplacian variance, entropy, texture complexity, and blockiness indicators support screening for blur, low-information content, noise, and compression artifacts. 
Layout and saliency checks identify heavily partitioned compositions and references in which the target product is insufficiently prominent. 
OCR-based text density and text--subject overlap help identify text that obscures the product.
The objective is to preserve informative promotional text while removing severe interference, not to eliminate text-bearing images. A binary classifier using CLIP~\cite{clip} and DINOv3~\cite{simeoni2025dinov3} features supplements these checks for image-level assessment.

\begin{figure}[t]
  \centering
  \includegraphics[pagebox=cropbox,width=0.95\linewidth]{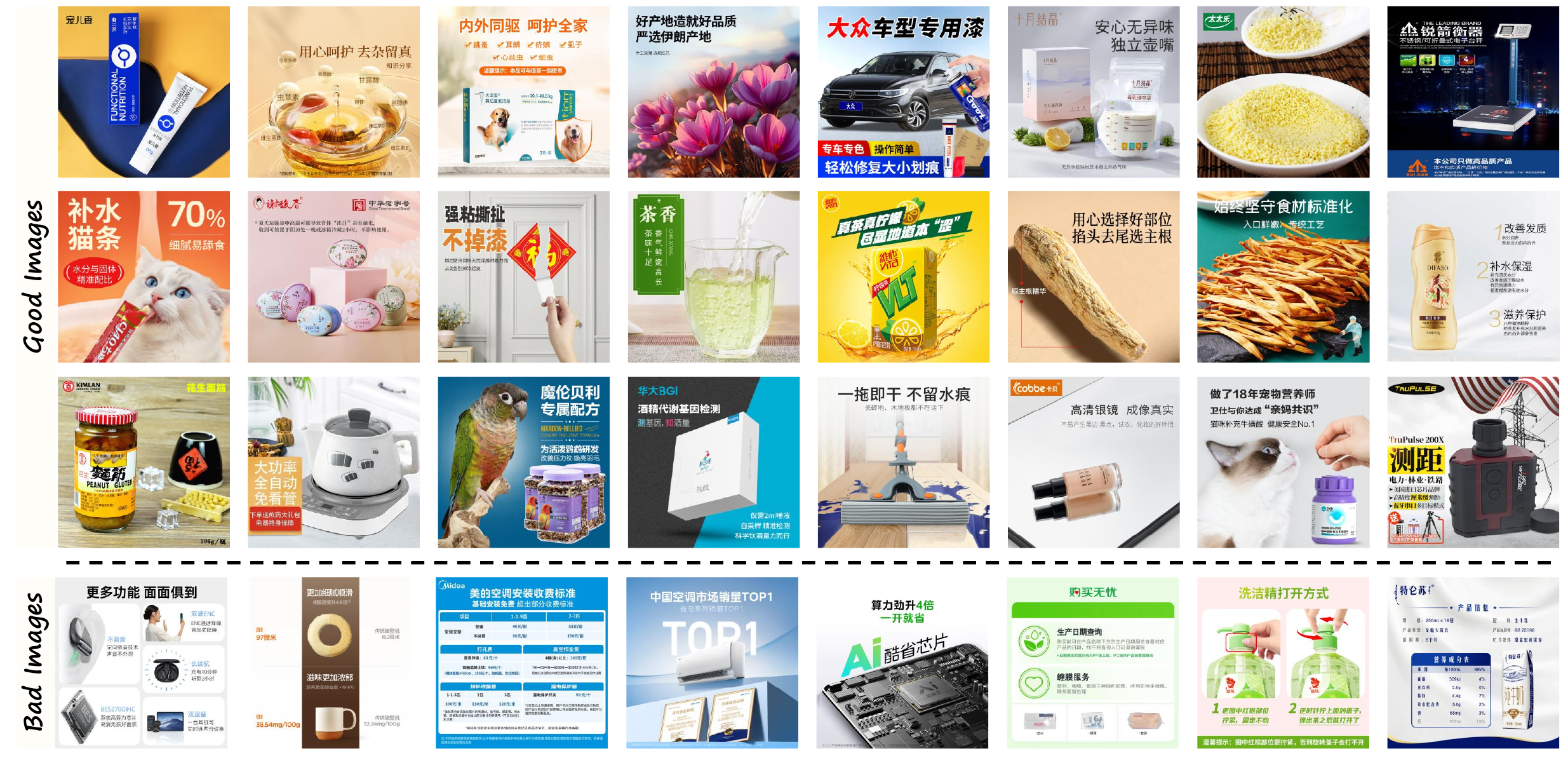}
  \caption{Examples of retained and excluded reference images in EVF-32K construction. The upper panels, labeled \emph{Good Images}, show retained product-centered presentations. The lower panels, labeled \emph{Bad Images}, show excluded references with dense specifications, instructional content, or multi-panel layouts. These labels indicate suitability for the present task.}
  \label{fig:dataset_reference_filtering}
\end{figure}

Fig.~\ref{fig:dataset_reference_filtering} shows representative retained and excluded images. Retained examples present recognizable products in compositions suited to dynamic presentation. Excluded examples include dense specifications, instructional layouts, and multi-panel compositions. 
Their exclusion concerns suitability as generation references rather than a universal judgment of image quality.

\begin{figure}[h]
  \centering
  \includegraphics[width=0.85\linewidth]{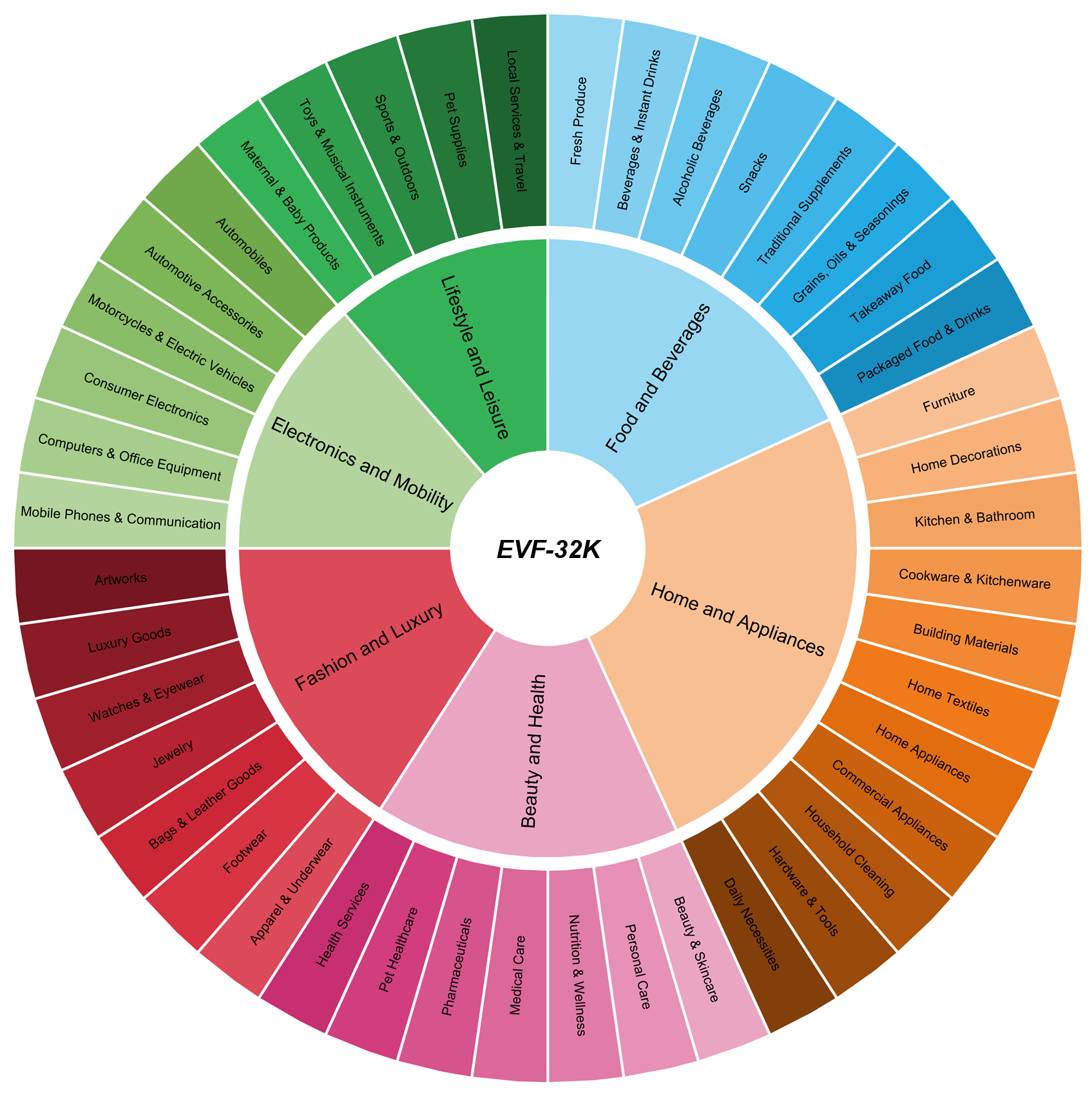}
  \caption{Product-category taxonomy of EVF-32K. The inner ring shows six product domains, and the outer ring lists their 44 fine-grained categories. Sector sizes are schematic and do not represent sample proportions.}
  \label{fig:dataset_product_taxonomy}
\end{figure}

\subsubsection{\textbf{Prompt Construction}}
We use a multi-modal GPT model~\cite{openai20205gpt5} to construct generation prompts from the curated images and structured task instructions. The prompts describe how the scene should evolve, including product movement, lighting changes, or surrounding effects. The instruction templates emphasize plausible dynamics, product visibility, stable promotional text, and a fixed viewpoint where appropriate. This stage connects the reference appearance to a concrete dynamic presentation while keeping the intended effect compatible with the product context.

\subsubsection{\textbf{Image-to-Video Generation}}
The image--prompt pairs are processed by image-to-video systems, including Kling, Jimeng, and Wan~\cite{kwai2025keling,byte2025jimeng,wan2025wan}, to produce candidate clips. Multiple systems and varied generation configurations provide different realizations of product motion and visual effects. Manual review supports the removal of severe structural inconsistencies and product distortions. These systems serve as data-generation sources; the evaluation of baseline and domain-adapted models is addressed in the experiments.

\subsubsection{\textbf{Video Filtering}}
Candidate videos are screened using motion-strength statistics, text-detection stability, visual sharpness, and product--frame similarity. These signals target clear failures such as near-static output, collapsed text, and severe changes in product appearance. They are applied as separate diagnostic checks rather than combined into an overall benchmark ranking score. The filtering stage selects usable product-centered examples while retaining variation in effects and motion. As filtering and evaluation examine related visual properties, the retained corpus is a curated collection rather than an unbiased sample of all outputs from its generating systems.

\subsubsection{\textbf{Metadata Augmentation}}
In the final stage, we augment each retained video with descriptive, structural, and quality-related metadata. Qwen2.5-VL~\cite{Qwen25VL} generates a caption summarizing the product appearance, scene content, and motion patterns observed in the video. We also record structural attributes, including resolution and frame count, together with quantitative quality indicators. These annotations enrich the dataset with information about both video content and technical characteristics, supporting content-based retrieval and analysis of product presentation and dynamic effects.

\subsection{Dataset Statistics}
\label{sec:dataset_statistics}
EVF-32K contains over $32{,}000$ examples with a nominal video resolution of $960\times960$ and an average duration of approximately 5 seconds, as summarized in Table~\ref{tab:dataset_comparison}. Its six product domains are \emph{Food \& Beverages} (30.1\%), \emph{Fashion \& Luxury} (24.3\%), \emph{Home \& Appliances} (18.8\%), \emph{Lifestyle \& Leisure} (9.7\%), \emph{Electronics \& Mobility} (8.6\%), and \emph{Beauty \& Health} (8.6\%). The taxonomy further comprises 44 fine-grained categories, as illustrated in Fig.~\ref{fig:dataset_product_taxonomy}. This coverage spans different product geometries, materials, and presentation contexts, with food and fashion accounting for the largest portions of the dataset.

\begin{figure*}[ht]
  \centering
  \includegraphics[pagebox=cropbox,width=0.95\textwidth]{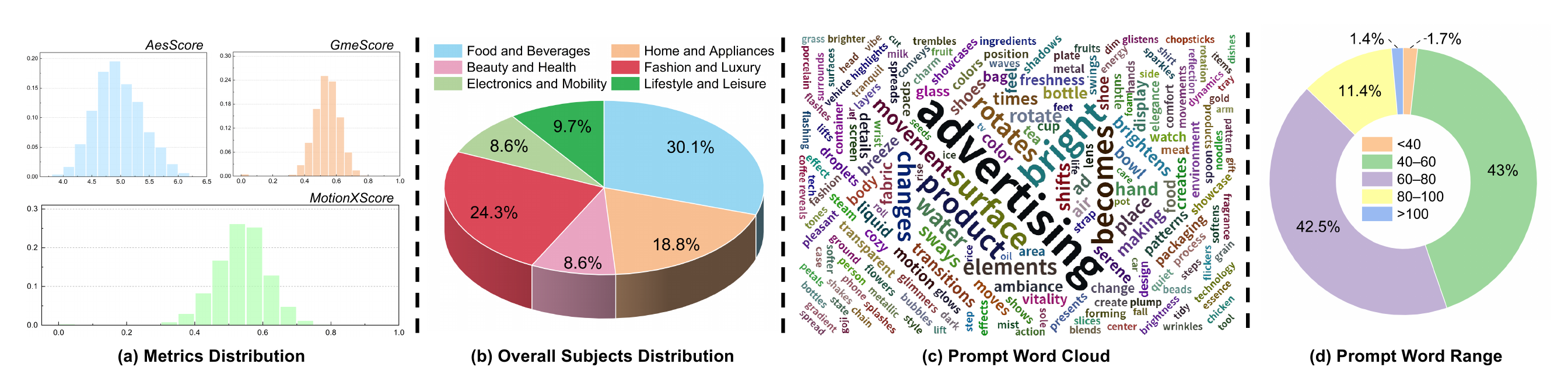}
  \caption{Statistics of EVF-32K: (a) distributions of AesScore, GmeScore, and MotionXScore; (b) product-domain distribution; (c) prompt word cloud; and (d) prompt-length distribution.}
  \label{fig:dataset_statistics_overview}
\end{figure*}

Fig.~\ref{fig:dataset_statistics_overview} summarizes score distributions, product-domain coverage, and textual characteristics. The score panel reports AesScore, GmeScore, and MotionXScore, illustrating variation in aesthetic quality, prompt relevance, and motion among the curated clips. The domain panel shows coverage across the six product domains. The word cloud highlights product attributes, movements, and visual effects, while the reported prompt-length distribution places most descriptions in the 40--80-word range. Together, these views characterize the corpus and the diversity of its product presentations.

\section{Experiments}
\label{sec:experiments}

Our experiments examine model performance across the six dimensions of EVF-Bench, the effect of adaptation using EVF-32K, and the correspondence between automatic measurements and human preferences. We combine quantitative comparisons with representative video examples and an analysis of the motion-persistence component.

\subsection{Evaluation Setups}
\label{sec:experimental_setups}
\subsubsection{\textbf{Evaluated systems}}
We compare four commercial systems, Taobao Xingchen~\cite{taobao2025xingchen}, Hailuo 02~\cite{minimax2025hailuo}, Seedance 1.0 Mini~\cite{byte2025jimeng}, and Kling 2.1 Std~\cite{kwai2025keling}, with four open-source models: LTX-Video-13B~\cite{ltxvideo}, HunyuanVideo~\cite{hunyuanvideo}, StepVideo~\cite{stepvideo}, and Wan2.2-A14B~\cite{wan2025wan}. To investigate task-aligned adaptation, we additionally evaluate LTX-EVF and Wan2.2-EVF, obtained by adapting the corresponding LTX and Wan backbones using EVF-32K. These ten configurations cover commercial systems, released research models, and adapted counterparts.

\subsubsection{\textbf{Generation conditions}}
The evaluation cases supply product reference images and prompts describing the requested presentation. Commercial systems are accessed through their official services, and open-source models use released checkpoints and recommended inference settings. The reference image and intended description provide the common basis for comparison. For Taobao Xingchen, the source protocol records a prompt-length restriction; descriptions exceeding the limit are shortened and manually refined to preserve their main content.

\subsubsection{\textbf{Domain adaptation}}
The adaptation recipe uses LoRA~\cite{Lora} modules in transformer attention layers with rank 64. Parameter-efficient transfer of this kind has also proven effective in other vision tasks that must bridge a domain gap, including selective mixture-of-experts designs for robust stereo matching in the wild~\cite{wang2025stereomoe}. Text encoders remain frozen, while adaptation is applied to the video backbone. Training uses AdamW with a learning rate of $10^{-4}$ and batch size 16. Each adapted model is compared with its original backbone using the same evaluation dimensions.

\subsubsection{\textbf{Implementation Details}}
\label{sec:experimental_implementation}
Open-source backbones are evaluated on an 8$\times$NVIDIA A100 GPU cluster, while commercial systems are accessed through their official services. All evaluation metrics are computed from the returned videos under the settings below.
For TSScore, the evaluation configuration uses $N_{\rm txt}=32$ uniformly sampled frames and a clarity mixing coefficient of $\lambda_c=0.7$. For MotionXScore, the temporal smoothing coefficient is $\alpha_m=0.5$, and the minimum active-run duration is $L_{\min}=0.3$ seconds. For CRScore, 16 frames are uniformly sampled and $N_{\rm cr}=3$ judgments are averaged per video. These settings instantiate the definitions in Section~\ref{sec:benchmark_metrics}. Results are reported separately for all six dimensions, with higher values indicating better performance within each metric.

\subsection{Quantitative Evaluation}
\label{sec:experimental_quantitative}

\begin{figure*}[!p]
  \centering
  \includegraphics[width=\linewidth]{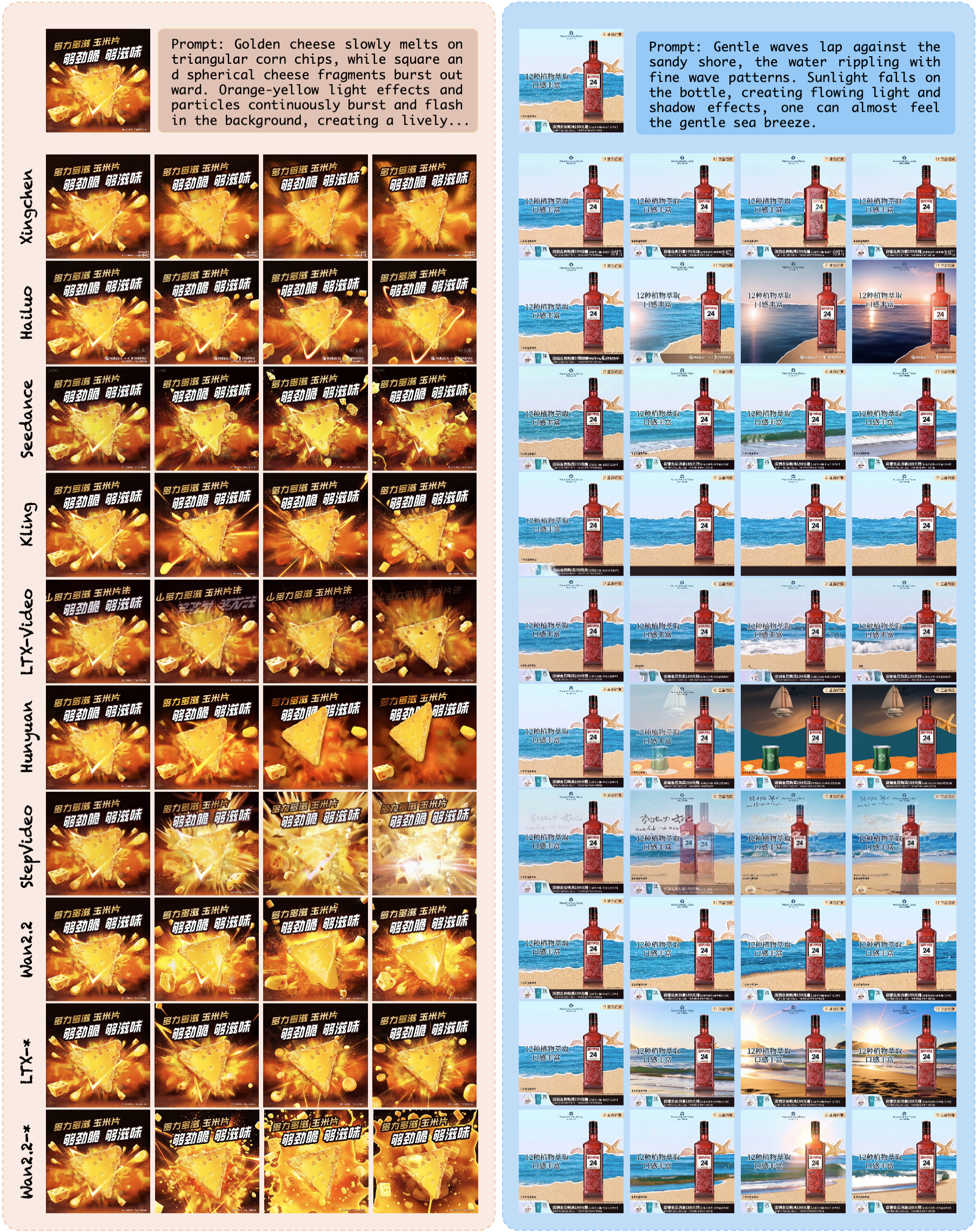}
  \caption{Qualitative comparison of commercial systems, open-source backbones, and adapted variants on two EVF cases. The top panels contain the reference images and prompts; each row below shows sampled frames from one configuration. LTX-* and Wan2.2-* denote LTX-EVF and Wan2.2-EVF, respectively.}
  \label{fig:experimental_qualitative}
\end{figure*}

Table~\ref{tab:quantitative} reports the six-dimensional performance of the evaluated configurations. We analyze the scores by dimension and compare each adapted variant with its backbone, rather than averaging the columns into an overall ranking.

\begin{table*}[t]
  \caption{Quantitative comparison on EVF-Bench across six evaluation dimensions. Higher scores indicate better performance within each column. Values retain the percentage-style reporting scale of the source results. Best values are bold and second-best values are underlined.}
  \label{tab:quantitative}
  \centering\small
  \setlength{\tabcolsep}{3pt}
  \renewcommand{\arraystretch}{1.08}
  \begin{tabular*}{0.95\textwidth}{@{\extracolsep{\fill}}llrrrrrr@{}}
    \toprule
    \textbf{Type} & \textbf{Method} & \textbf{AesScore}$\uparrow$ & \textbf{GmeScore}$\uparrow$ & \textbf{MotionX}$\uparrow$ & \textbf{ProductScore}$\uparrow$ & \textbf{TSScore}$\uparrow$ & \textbf{CRScore}$\uparrow$ \\
    \midrule
    Commercial & Taobao Xingchen~\cite{taobao2025xingchen} & 26.90\% & \textbf{59.03\%} & 54.33\% & 47.31\% & 78.26\% & 60.31\% \\
    & Hailuo 02~\cite{minimax2025hailuo} & 27.11\% & 58.32\% & \underline{55.13\%} & 48.42\% & 77.36\% & 65.14\% \\
    & Seedance 1.0 Mini~\cite{byte2025jimeng} & 27.52\% & 57.17\% & 53.02\% & 45.16\% & 75.19\% & 62.00\% \\
    & Kling 2.1 Std~\cite{kwai2025keling} & \textbf{31.76\%} & 58.14\% & \textbf{56.58\%} & \underline{50.64\%} & \textbf{80.80\%} & \underline{66.25\%} \\
    \midrule
    Open-source & LTX-Video-13B~\cite{ltxvideo} & 20.20\% & 56.83\% & 49.11\% & 33.79\% & 64.70\% & 54.87\% \\
    & HunyuanVideo~\cite{hunyuanvideo} & 28.16\% & 54.50\% & 48.01\% & 39.31\% & 69.76\% & 54.19\% \\
    & StepVideo~\cite{stepvideo} & 20.68\% & 53.22\% & 48.31\% & 31.05\% & 63.84\% & 52.50\% \\
    & Wan2.2-A14B~\cite{wan2025wan} & 29.05\% & 56.78\% & 48.34\% & 41.86\% & 72.82\% & 60.65\% \\
    \midrule
    Adapted & LTX-EVF & 27.93\% & 58.02\% & 50.01\% & 40.06\% & 70.95\% & 57.10\% \\
    & Wan2.2-EVF & \underline{31.47\%} & \underline{58.90\%} & 54.05\% & \textbf{50.71\%} & \underline{78.42\%} & \textbf{67.23\%} \\
    \bottomrule
  \end{tabular*}
\end{table*}

\begin{figure*}[ht]
  \centering
  \includegraphics[pagebox=cropbox,width=0.95\textwidth]{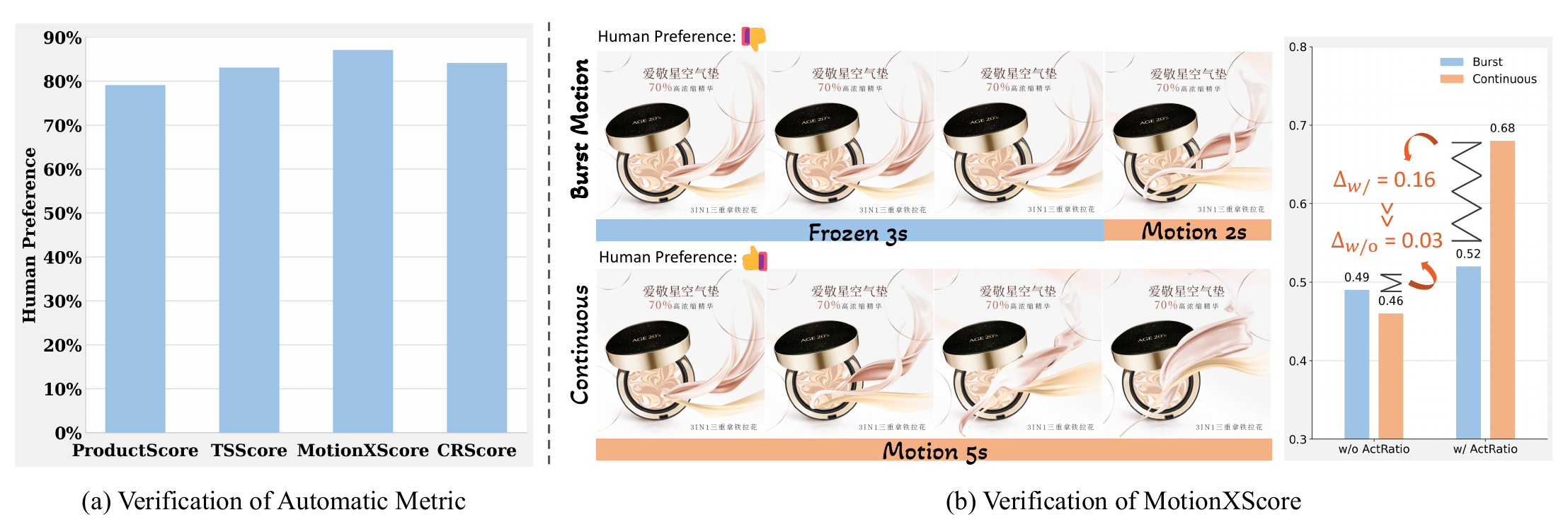}
  \caption{Human evaluation and motion analysis. (a) Reported agreement between automatic metrics and human preferences for four EVF-specific dimensions. (b) A case comparing burst and sustained motion, with scores shown before and after adding the activation-ratio component. The illustrated human preference favors sustained motion.}
  \label{fig:experimental_human_metrics}
\end{figure*}

\subsubsection{\textbf{Complementary strengths}}
No evaluated configuration leads all six dimensions. Kling 2.1 Std obtains the highest AesScore (31.76), MotionXScore (56.58), and TSScore (80.80), whereas Taobao Xingchen achieves the highest GmeScore (59.03). Wan2.2-EVF leads in ProductScore (50.71) and CRScore (67.23). Thus, the highest product consistency score does not coincide with the highest aesthetic or motion score, and the strongest prompt relevance does not imply the strongest text stability. These different leaders illustrate the value of reporting the individual dimensions.

\subsubsection{\textbf{Product and text preservation}}
The commercial systems score higher than the four unadapted open-source models in both ProductScore and TSScore. ProductScore ranges from 45.16 to 50.64 for the commercial systems and from 31.05 to 41.86 for the open-source backbones; the corresponding TSScore ranges are 75.19--80.80 and 63.84--72.82. This difference is not uniform across all dimensions: HunyuanVideo has an AesScore of 28.16, exceeding three of the commercial systems, while its ProductScore and TSScore remain lower. Appearance quality and reliable product information therefore show different performance patterns under the evaluated conditions.

\subsubsection{\textbf{Effect of domain adaptation}}
Both adapted variants improve over their respective backbones across all six reported scores. LTX-EVF gains 7.73 points in AesScore, 6.27 in ProductScore, and 6.25 in TSScore, while its MotionXScore gain is 0.90. Wan2.2-EVF improves ProductScore by 8.85 points, CRScore by 6.58, MotionXScore by 5.71, and TSScore by 5.60. These differences show that adaptation can improve several EVF requirements together, although the magnitude of improvement varies by model and dimension. The paired comparisons support further investigation of task-aligned adaptation using EVF-32K.

\subsection{Qualitative Evaluation}
\label{sec:experimental_qualitative}
Fig.~\ref{fig:experimental_qualitative} compares sampled frames from two EVF cases. The first requests melting cheese, corn chips, and energetic surrounding effects; the second requests a product bottle against a beach scene with waves and changing illumination. The cases expose different challenges in product appearance, promotional text, and dynamic presentation.

In the food case, the LTX-Video output loses prominent promotional text in later displayed frames, while StepVideo introduces large scene changes and extensive effects. The adapted LTX and Wan outputs retain visible text while producing dynamic variations around the product. Because melting is explicitly requested, appearance changes must be interpreted in relation to the prompt rather than treated as errors solely because frames differ from the reference image.

In the bottle case, several commercial outputs keep the bottle recognizable with relatively small changes across the sampled frames. HunyuanVideo changes the beach context to an indoor presentation in the later frames, and StepVideo exhibits substantial variation in the rendered text and product appearance. These examples illustrate why scene-level relevance, product consistency, and text stability are evaluated separately. The visual comparisons help interpret the quantitative results without estimating failure frequencies or full-video temporal quality from selected frames alone.

\subsection{Human Evaluation and Metric Analysis}
\label{sec:experimental_human_metrics}
\subsubsection{\textbf{Agreement with human preference}}
The human evaluation involves 30 volunteers, each comparing 30 video pairs generated from the same case by different models or configurations. Participants select the preferred output in terms of product consistency, text stability, motion, and creative rationality. Preferences are aggregated by majority vote, and the resulting pairwise choices are compared with the rankings assigned by the corresponding automatic metrics. Fig.~\ref{fig:experimental_human_metrics}(a) reports agreement across the four EVF-specific dimensions. Large multimodal models have recently been used to predict perceptual video quality ratings directly~\cite{ge2025lmmvqa}; the EVF-specific metrics instead keep each dimension separate and are compared here with pairwise preferences. The reported agreement supports the use of these metrics as diagnostic proxies for perceptual assessment in the sampled comparisons.

\subsubsection{\textbf{Motion-persistence case study}}
Fig.~\ref{fig:experimental_human_metrics}(b) compares two five-second clips: one remains frozen for three seconds and exhibits motion for two seconds, while the other sustains motion across the clip. Without the activation-ratio component, the scores are 0.49 for burst motion and 0.46 for sustained motion. After including the component, the scores become 0.52 and 0.68, respectively, changing the ordering to match the illustrated human preference. The absolute score gap increases from 0.03 to 0.16. This case demonstrates how activation modeling adds information about temporal persistence beyond the inherited motion components; it is an illustrative comparison rather than a benchmark-wide ablation.

\section{Conclusion}
\label{sec:conclusion}
This work introduces EVF-Bench, a dedicated benchmark for evaluating e-commerce visual effects in video generation. It combines visual quality and text relevance with motion, product consistency, text stability, and creative rationality, providing a six-dimensional view of product-centered video presentations. To support research and model adaptation in this setting, we construct EVF-32K, a supporting dataset of over $32{,}000$ curated image--text--video triples.

Evaluation of representative video generation systems and their domain-adapted variants reveals complementary strengths, with no evaluated configuration leading all six dimensions. Both adapted variants achieve higher reported scores than their respective backbones, although the gains differ across models and dimensions. These findings highlight the value of task-oriented evaluation for distinguishing visual appeal, reliable product information, and dynamic presentation. Together, EVF-Bench and EVF-32K provide an evaluation framework and a complementary data resource for studying and improving e-commerce video generation.

\bibliographystyle{IEEEtran}
\bibliography{references}


\end{document}